\documentclass[runningheads]{llncs}
\usepackage{marvosym}
\usepackage{graphicx}
\usepackage{amsfonts} 
\usepackage{amsmath}
\usepackage{multirow}
\usepackage{booktabs}
\usepackage{colortbl}
\usepackage[table]{xcolor}
\usepackage[normalem]{ulem}
\useunder{\uline}{\ul}{}

\usepackage{bbding}
\usepackage{orcidlink}

\begin{document}

\title{WaMaIR: Image Restoration via Multiscale Wavelet Convolutions and Mamba-based Channel Modeling with Texture Enhancement}
\titlerunning{WaMaIR}

\author{
  Shengyu Zhu\textsuperscript{*} \quad
  Fan\textsuperscript{*, \Letter} \quad
  Fuxuan Zhang
}
\authorrunning{S.~Zhu et al.}
\institute{
  Harbin Engineering University \\
  % \email{\{zsy\_bg, fancongyi, fuxuan\}@hrbeu.edu.cn}%
}
\renewcommand{\thefootnote}{\fnsymbol{footnote}}
\footnotetext{* Equal contribution}
\footnotetext{\Letter\ Corresponding author}
\renewcommand{\thefootnote}{\arabic{footnote}}

\maketitle              % typeset the header of the contribution
\begin{abstract}

Image restoration is a fundamental and challenging task in computer vision, where CNN-based frameworks demonstrate significant computational efficiency. However, previous CNN-based methods often face challenges in adequately restoring fine texture details, which are limited by the small receptive field of CNN structures and the lack of channel feature modeling. In this paper, we propose WaMaIR, which is a novel framework with a large receptive field for image perception and improves the reconstruction of texture details in restored images. Specifically, we introduce the Global Multiscale Wavelet Transform Convolutions (GMWTConvs) for expandding the receptive field to extract image features, preserving and enriching texture features in model inputs. Meanwhile, we propose the Mamba-Based Channel-Aware Module (MCAM), explicitly designed to capture long-range dependencies within feature channels, which enhancing the model sensitivity to color, edges, and texture information. Additionally, we propose Multiscale Texture Enhancement Loss (MTELoss) for image restoration to guide the model in preserving detailed texture structures effectively. Extensive experiments confirm that WaMaIR outperforms state-of-the-art methods, achieving better image restoration and efficient computational performance of the model.

\keywords{Image restoration \and Texture details \and Wavelet transform \and Channel modeling.}
\end{abstract}

\section{Introduction}

Image restoration, as one of the important tasks in the field of computer vision, aims to recover clean latent images from degraded observations, such as dehazing, deraining and desnowing. It has wide applications in remote sensing, autonomous driving, and medical imaging~\cite{wei2024optical,lin2025jarvisir,yang2024all}. 
However, preserving detailed textures during image restoration remains challenging, as it necessitates effectively capturing and utilizing complex texture patterns and their spatial distribution.

Traditional image restoration algorithms, which rely heavily on handcrafted features to constrain the solution space, tend to be impractical for real-world scenarios~\cite{zhang2022deepimagedeblurringsurvey}.
With the rapid advancement of deep learning, various modern architectures, such as convolutional neural network (CNN)~\cite{lecun2002gradient} and Transformer~\cite{vaswani2017attention}, have been widely adopted for image restoration.

CNN-based methods, such as ConvIR~\cite{cui2024revitalizing}, DehazeNet~\cite{cai2016dehazenet}, DesnowNet~\cite{liu2018desnownet}, offer computational efficiency but suffer from limited receptive fields, restricting their ability to model long-range dependencies, thus hindering their performance in image restoration.
Transformer-based methods, including AST~\cite{zhou2024adapt}, UHDformer~\cite{wang2024correlation}, GridFormer~\cite{wang2024gridformer}, leverage attention mechanisms to effectively model long-range dependencies; however, they incur a significant drawback of quadratic computational complexity.

To this end, we propose WaMaIR, a novel framework integrating multiscale wavelet convolutions, Mamba-based channel modeling, and a multiscale texture enhancement loss to achieve superior preservation of texture details and enhanced image restoration performance. 
Unlike conventional designs, our framework leverages convolutional spatial encoding in combination with Mamba-based sequence modeling to enhance image restoration performance.
Specifically, WaMaIR introduces Global Multiscale Wavelet Transform Convolutions (GMWTConvs), a globally-aware receptive field module that leverages wavelet transforms to extract multiscale high-frequency and low-frequency image features.
GMWTConvs effectively expands the receptive field, enabling the modeling of long-range features, thereby improving texture detail preservation and overall restoration quality.

Furthermore, to overcome the limitations of previous methods that inadequately capture channel-wise features, we propose the Mamba-Based Channel-Aware Module (MCAM). Motivated by recent advancement demonstrating the efficiency of Mamba architectures in modeling long-range dependencies~\cite{gu2023mamba}, MCAM explicitly encodes global dependencies across feature channels by integrating global pooling operations and the Mamba sequence model. This design enhances model sensitivity to color, edge, and texture attributes, leading to improved restoration outcomes.

To further enhance texture representation, we introduce the Multiscale Texture Enhancement Loss (MTELoss). Traditional loss functions often overemphasize smoothness or global consistency~\cite{cui2023irnext,cui2024revitalizing}, leading to insufficient preservation of fine textures. In contrast, our MTELoss explicitly targets the multiscale texture details extracted via wavelet decomposition, effectively guiding the model to precisely reconstruct and enhance intricate details that high-frequency and low-frequency domains.

Extensive experiments on four datasets across dehazing, deraining, and desnowing tasks demonstrate that WaMaIR achieves state-of-the-art performance with superior detail preservation and overall image restoration quality.

Our main contributions can be summarized as follows: 
(1) We propose WaMaIR, a novel framework specifically designed for effectively preserving texture details in image restoration.
(2) We introduce Global Multiscale Wavelet Transform Convolutions (GMWTConvs) to expand receptive fields explicitly for capture multiscale texture features.
(3) We propose Mamba-based Channel-Aware Module (MCAM) to efficiently capture long-range channel dependencies with linear computational complexity.
(4) We introduce Multiscale Texture Enhancement Loss (MTELoss) effectively guiding the model to precisely reconstruct and enhance reconstructing images texture details.

\section{Related Work}
\subsection{Image Restoration}

Image restoration aims to recover clean latent images from degraded observations, representing a classic inverse problem. It can be primarily categorized into two types: methods based on physical models and those relying on deep learning.  
Traditional physical model-based approaches were commonly used, such as the atmospheric scattering model (for image dehazing) \cite{he2010single} and handcrafted priors \cite{buades2005non}. While these methods are effective in specific scenarios, their generalization capability is limited, making them less suitable for handling complex degradations.

In recent years, deep learning-based approaches have gained prominence in image restoration tasks, including Transformer-based methods\cite{zamir2022restormer,wang2022uformer}. Among them, CNN-based architectures\cite{cui2023irnext,cui2024revitalizing} are widely adopted due to their strong performance in local feature extraction  \cite{cai2016dehazenet,liu2018desnownet}. However, they struggle to model long-range dependencies \cite{zhang2017beyond}. Although increasing the kernel size can expand the receptive field—as demonstrated in \cite{ding2022scaling}, where larger convolutional kernels were explored—performance eventually saturates once the kernel reaches a certain size, with a quadratic impact on computational cost. Thus, balancing performance and receptive field remains an unresolved challenge in image restoration tasks.

\subsection{State Space Models}

The recently proposed Mamba \cite{gu2023mamba} further advances the field with dynamic parameters, hardware optimization, and architectural simplification, demonstrating outstanding performance in computer vision, natural language processing, and other domains. 
In the field of image restoration, Mamba has also shown great potential. VmambaIR \cite{shi2025vmambair} employs a U-Net architecture stacked with Omnidirectional Selective Scan (OSS) blocks, which consist of OSS modules and Efficient Feed-Forward Networks (EFFN). MambaIR \cite{guo2024mambair} enhances standard Mamba by incorporating convolution and channel attention. CU-Mamba \cite{deng2024cu} introduces a Channel SSM module within the U-Net architecture to compress and reconstruct features along the channel dimension. This study further explores Mamba’s ability to capture long-range dependencies in the channel dimension, aiming to expand the receptive field in image restoration tasks while maintaining linear complexity.

\section{Method}

In this section, we first describe the overall pipeline of WaMaIR. Then we present the core components of WaMaIR:  Global Multiscale Wavelet Transform Convolutions (GMWTConvs) and Mamba-based channel-aware module (MCAM). The multiscale texture enhancement loss (MTELoss) is introduced in the final part.

\begin{figure}[t]
    \centering
    \includegraphics[width=1\linewidth]{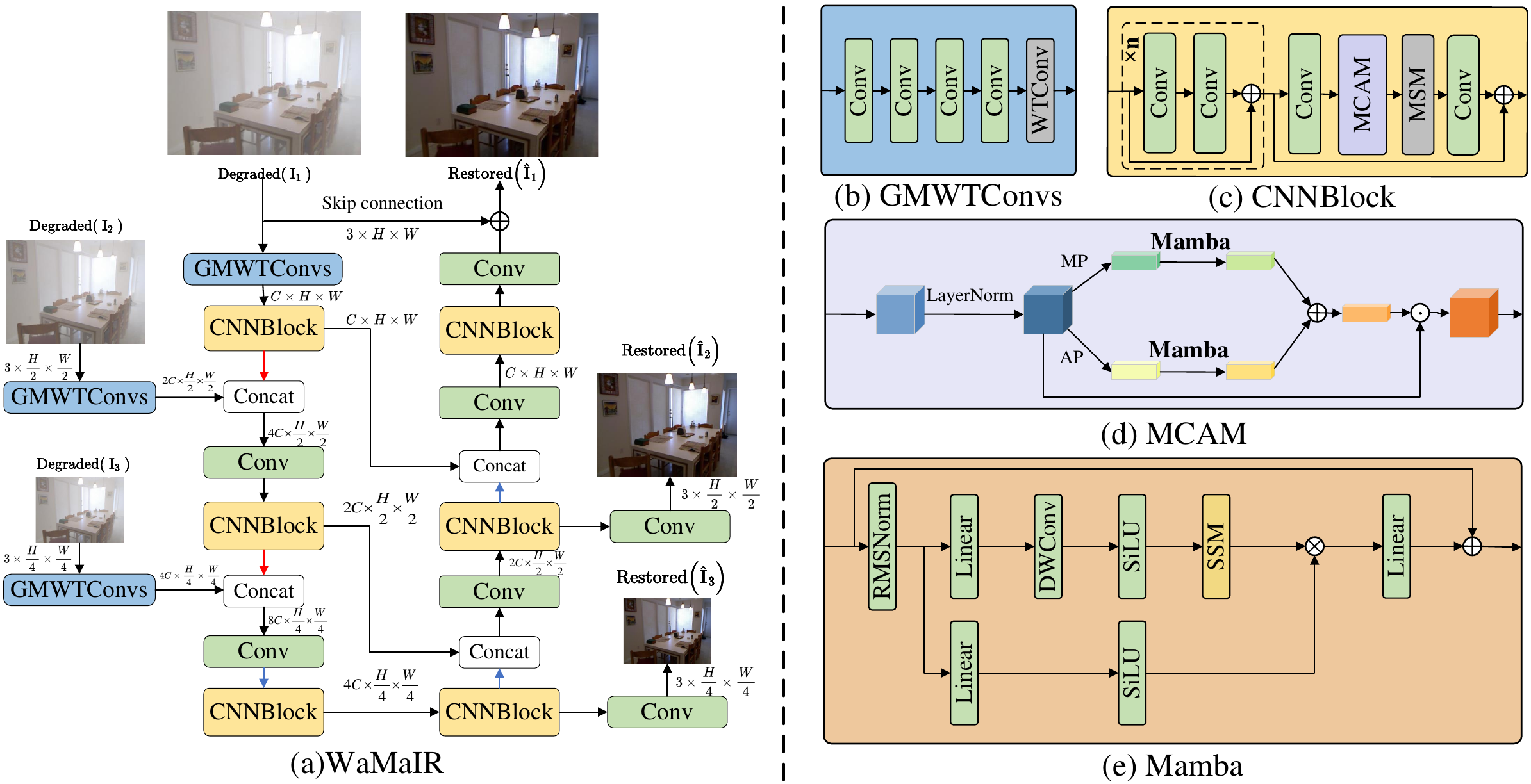}
    \caption{The architecture of our WaMaIR. (a) WaMaIR adopts a U-shaped structure, consisting of several CNN Blocks and three GMWTConvs, and employs a multi-input multi-output strategy. (b) GMWTConvs first extract shallow features through four convolutional layers, then use WTConv to capture features in different frequency domains. (c) CNNBlock contains multiple residual connections, with our proposed MCAM added at the end. (d) MCAM enhances Mamba's channel feature extraction capability through AP and MP. (e) The structure of Mamba.}
    \label{fig:pipeline}
    \vspace{-0.5cm}
\end{figure}

\subsection{Overall Architecture}

As shown in Figure~\ref{fig:pipeline} (a), the proposed WaMaIR adopts a U-shaped architecture for image restoration.
Specifically, given a degraded image $\mathbf{I} \in \mathbb{R}^{3 \times H \times W}$, we first utilize our proposed global multiscale wavelet transform convolutions (GMWTConvs), as shown in Figure~\ref{fig:pipeline} (b) to generate shallow texture features with the size of $C \times H \times W$, where $C$ denotes the number of feature channels and $H \times W$ represents feature spatial locations.
Then the shallow texture features through three CNNBlocks to extract high-level features. Each CNNBlock contains multiple residual convolution blocks with our mamba-based channel-aware module (MCAM) then follow a multiscale module (MSM) from ConvIR~\cite{cui2024revitalizing} in the last one, as shown in Figure~\ref{fig:pipeline} (c).

During this process the spatial resolution is reduced, while the channels are expended, multiple downsampled degraded images are extract feature via the GMWTConvs module. Then downsampled degraded images are merged into the main path via concatenation and adjust the number of channels thourgh a 3$\times$3 convolution. These enhanced features facilitate richer feature representations, greatly benefiting subsequent image reconstruction. Next the high-level are fed into another three CNNBlocks to restore the features. Following previous methods~\cite{cui2023irnext,cui2024revitalizing}, multi-output strategy is adopted, where the low-resolution predicted image is generated using a 3$\times$3 convolution and image-level skip connection. The multiscale predicted images are fed into our proposed Multiscale Texture Enhancement Loss (MTELoss) for loss computation.
Next, we describe in detail the proposed GMWTConvs, MCAM and MTELoss.

\subsection{Global Multiscale Wavelet Transform Convolutions}

To better preserve texture details and enhance image restoration performance, we propose the Global Multiscale Wavelet Transform Convolutions (GMWTConvs) module, which leverages wavelet transforms to enhance the extraction of texture information and effectively expand receptive fields. The architecture of GMWTConvs module is shown in the Figure~\ref{fig:pipeline} (b). Unlike existing methods that only use multilayer ordinary convolution~\cite{cui2023irnext,cui2024revitalizing}, GMWTConvs incorporates wavelet transforms to enhance the extraction of texture information, significantly enlarging the receptive field to better capture degraded image features.

Inspired by WTConv~\cite{finder2024wavelet}, which effectively achieves large receptive fields through wavelet-based convolutions, we propose an enhanced structure tailored specifically for image restoration tasks. Traditional convolutional neural networks typically require enlarging kernel sizes to expand receptive fields, inevitably causing quadratic parameter growth. Additionally, standard convolutions predominantly capture high-frequency features, making them suboptimal for restoration scenarios such as dehazing, where the degradation usually resides in low-frequency components. To overcome these challenges, we introduce GMWTConvs, which multiple downsampled degraded image through it extract different frequency feature.

\begin{figure}
    \centering
    \includegraphics[width=\linewidth]{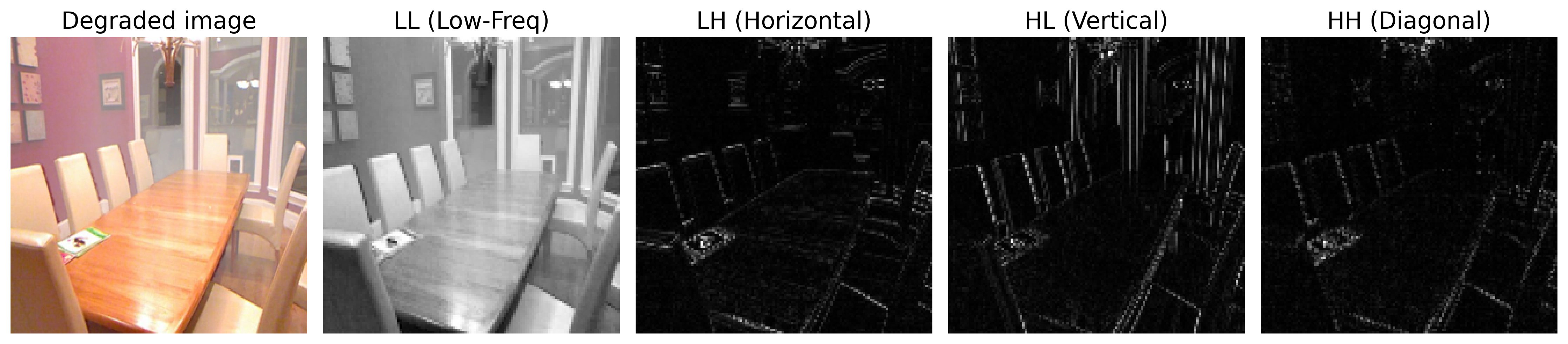}
    \caption{Visualization results of degraded image after wavelet transform. In our work, we employ the Haar wavelet transform, which more effectively captures image textures after decomposition. Specifically, LL represents the overall contour of the image, LH reflects vertical edge features, HL reflects horizontal edge features, and HH represents diagonal details in the image.}
    \label{fig:wavelet}
\end{figure} 

In our work, we employ the Haar wavelet transform and find that it enables a more effective representation of image texture features, as shown in Figure~\ref{fig:wavelet}. The transformed image reveals four sub-bands: LL represents the overall contour and basic structure of the image, obtained by applying low-pass filtering along both rows and columns; LH captures horizontal details and reflects vertical edge features, resulting from low-pass filtering along rows and high-pass filtering along columns; HL corresponds to vertical details and reflects horizontal edge features, derived from high-pass filtering along rows and low-pass filtering along columns; HH contains diagonal details of the image, including high-frequency noise and textures, obtained by applying high-pass filtering in both directions.

Specifically, GMWTConvs consists of multiple layers of ordinary convolution and one layer of WTConv~\cite{finder2024wavelet}, as shown in Figure~\ref{fig:pipeline} (b).
Each layer of GMWTConvs extracts a different scale resolution $\mathrm{I_1} \in \mathbb{R}^{3 \times H \times W}$, $\mathrm{I_2} \in \mathbb{R}^{3 \times H \times W}$, and $\mathrm{I_3} \in \mathbb{R}^{3 \times H \times W}$, the process is processed by:
\begin{equation}
    \mathrm{WTConv}(X)=\mathrm{IWT}(\mathrm{Conv}(W,\mathrm{WT}(X))),
\end{equation}
where $X$ denotes the input feature, WT and IWT represent the wavelet transform and inverse wavelet transform, $W$ denotes the weight tensor of the convolution kernal. Our GMWTConvs is as follows:
\begin{equation}
    \mathrm{GMWTConvs}=\mathrm{WTConv}(\mathrm{Conv^{(4)}}(X)),
\end{equation}
where $\mathrm{Conv^{(4)}}$ represents four convolution operations on the feature. This multiscale frequency domain decomposition enables the network to better capture both global structures and intricate textures simultaneously.

\subsection{Mamba-Based Channel-Aware Module}

To address the limitation of previous methods in adequately capturing channel-wise features, we propose the Mamba-Based Channel-Aware Module (MCAM), as shown in Figure~\ref{fig:pipeline}. Existing Mamba-based neural networks excel at capturing global context, they remain less sensitive to channel-wise information. 
Inspired by channel attention mechanisms \cite{hu2018squeeze}, we design the MCAM, a Mamba-based module that utilizes global pooling to learn channel representations.

The core of Mamba is the SSM mechanism, and the overall architecture of Mamba is shown in Figure~\ref{fig:pipeline}(e). SSM is a linear time-invariant structure that maps input signals to responses through hidden states. Specifically:
\begin{equation}
\begin{aligned}
    \mathbf{h}^{\prime}(t) & =\mathbf{Ah}(t)+\mathbf{B}u(t), \\
    \mathbf{y}(t) & =\mathbf{Ch}(t)+\mathbf{D}u(t),
\end{aligned}
\end{equation}
where, $\mathbf{A}\in\mathbb{R}^{N\times N}$ is the state transition matrix, $\mathbf{B}\in\mathbb{R}^{N\times 1}$ is the input weight matrix, $\mathbf{C}\in\mathbb{R}^{1\times N}$ is the output weight matrix, and $\mathbf{D}\in\mathbb{R}^{1}$ is the feedforward parameter.

Different from CU-Mamba \cite{deng2024cu}, which uses a single-path design, we employ a dual-path strategy to capture both low- and high-frequency spatial information—using Average Pooling and Max Pooling respectively. This design is particularly well-suited for image restoration tasks, which often require suppressing low-frequency artifacts while preserving high-frequency details. Specially, given a layer-normalized input image $X^{{H}\times{W}\times{C}}$, we first apply global average pooling (AP) and global max pooling (MP) along the spatial dimensions, obtaining two 1×1×C feature maps:
\begin{equation}
    X_\mathrm{AP}=\mathrm AP(X), X_\mathrm{MP}=\mathrm MP(X),
\end{equation}

where $X_\mathrm{AP}$ represents the globally average-pooled features and $X_\mathrm{MP}$ denotes the max-pooled features. These feature maps are then reshaped into a 1×C sequence format and processed via Mamba to learn channel-wise dependencies. The outputs are summed:
\begin{equation}
    X_{\mathrm{Channel}}=X \odot (\mathrm Mamba(X_\mathrm{AP})+\mathrm Mamba(X_\mathrm{MP})),
\end{equation}
where $X_{SSM}$ represents the learned channel attention weights. Finally, we multiply the input image with these weights and employ a residual connection to preserve the original structure:
\begin{equation}
    X_{\mathrm{MCAM}}=X+X_{\mathrm{Channel}}.
\end{equation}

This explicit modeling of channel-wise dependencies greatly improves the representation power of channel features, enhancing the network's sensitivity to texture variations.

\subsection{Multiscale Texture Enhancement Loss}

We introduce Multiscale Texture Enhancement Loss (MTELoss) to explicitly guide our model towards preserving detailed textures. Unlike conventional losses, which overly prioritize global smoothness~\cite{cui2023irnext,cui2024revitalizing}, our MTELoss integrates spatial, frequency, and wavelet decomposition losses to comprehensively enhance multiscale texture recovery in the restored images.

MTELoss combines three distinct loss components, with the first two defined as follows:
\begin{equation}
    \mathcal{L}_{\text{spatial}} = \sum_{i=1}^{3} \frac{1}{P_i} \left\| \hat{I}_i - Y_i \right\|_1
\end{equation}
\begin{equation}
    \mathcal{L}_{\text{frequency}} = \sum_{i=1}^{3} \frac{1}{S_i} \left\| \left[ R(\mathcal{F}(\hat{I}_i)), S(\mathcal{F}(\hat{I}_i)) \right] - \left[ R(\mathcal{F}(Y_i)), S(\mathcal{F}(Y_i)) \right] \right\|_1 ,
\end{equation}
where the index $i$ refers to the three different scale outputs in the model (typically corresponding to 1/4, 1/2, and full resolutions); $\hat{I}_i$ and $Y_i$ represent the predicted images and ground truth images respectively; $P_i$ and $S_i$ denote total normalization factors; the operator $[\cdot,\cdot]$ indicates concatenation; $R(\cdot)$ and $S(\cdot)$ correspond to the real and imaginary components of the Fast Fourier Transform. Then we design our wavelet loss for learning the texture feature:
\begin{equation}
    \mathcal{L}_{\text{wavelet}} = \sum_{i=1}^{3} \sum_{b\in\{\text{LL,LH,HL,HH}\}} \text{SSIM}(\mathcal{W}_b(\hat{I}_i), \mathcal{W}_b(Y_i))
\end{equation}

$\mathcal{W}_b(\cdot)$ represents the Haar wavelet decomposition for extracting subband b, $b\in\{\text{LL,LH,HL,HH}\}$, where LL represents the low-frequency component, while LH, HL, and HH represent the high-frequency components in the vertical, horizontal, and diagonal directions. The predicted images and ground truth images are then evaluated via SSIM (Structural Similarity Index Measure) to compute structural similarity.

To effectively integrate wavelet-based features, we employ a MTELoss strategy to train our WaMaIR model. The overall loss function is given by:
\begin{equation}
    \mathcal{L}_{\text{MTE}} = \mathcal{L}_{\text{spatial}} + \theta \cdot \mathcal{L}_{\text{frequency}} + \lambda \cdot \mathcal{L}_{\text{wavelet}}
\end{equation}
where we set $\theta=0.1$ and $\lambda=0.05$ in our implementation. For weight selection, we followed the strategy in ConvIR~\cite{cui2024revitalizing} for spatial-domain and frequency-domain losses, assigning them balanced contributions based on empirical validation. For the wavelet-domain loss, we introduced a carefully tuned coefficient to address inherent scale differences: the high-frequency subbands (e.g., HH) in the wavelet domain typically have values one to two orders of magnitude smaller than those in the spatial domain. A weight set too high can lead to artificial textures in the restored images.
To mitigate this, we empirically set the wavelet-domain loss weight to half that of the frequency-domain loss, achieving a favorable trade-off between detail preservation and artifact suppression in ablation studies.
Utilizing wavelet transform ensures that structural information and textures are consistently preserved across different frequency bands.

\section{Experiments}

We evaluate the effectiveness of WaMaIR against current mainstream methods across four datasets covering three image restoration tasks: dehazing, desnowing, and deraining. First, we introduce the experimental setup; then, we present WaMaIR's results on each dataset and compare them with state-of-the-art methods; finally, we conduct detailed ablation studies. The best results in the tables are highlighted in \textbf{bold}.

\subsection{Experimental Setup}

All experiments were conducted using a single RTX 3090 GPU for both training and testing WaMaIR. Unless otherwise specified, we employed the Adam optimizer with an initial learning rate of 1e-4 that was gradually reduced to 1e-6 through cosine annealing, along with random horizontal flipping for data augmentation. The model architecture consisted of 3 WaveletConv layers and used Mamba configuration with d\_state=32, d\_conv=4, and expand=2, while computational complexity was measured using 256×256 image patches. Considering varying complexity across different restoration tasks, we adopted task-specific hyperparameters. As illustrated in Figure~
\ref{fig:pipeline}(c), task-specific hyperparameters were set with n=3 for dehazing and desnowing tasks, and n=15 for deraining tasks. The synthetic datasets are widely adopted in image dehazing, deraining, and desnowing tasks due to the scarcity of high-quality real-world datasets. To ensure fair and reproducible comparisons, we followed the previous works, conducting both quantitative and qualitative experiments on five widely used benchmark datasets \cite{li2018benchmarking,zhang2020nighttime,zhang2019image,fu2017removing,liu2018desnownet} covering three different low-level vision scenarios.

\begin{table}[h]
\renewcommand{\arraystretch}{0.97}
\setlength{\tabcolsep}{5pt}
\caption{Image Dehazing Comparisions on the Synthetic SOTS \cite{li2018benchmarking} Dataset}
\label{tab:reside-indoor}
\centering
\begin{scriptsize}
\begin{sc}
\begin{tabular}{l|cc|c|c}
\toprule
    \textbf{Methods} & \textbf{PSNR} & \textbf{SSIM} & \textbf{Params/M} & \textbf{FLOPs/G} \\
    \midrule
    DeHamer \cite{guo2022image} & 36.63 & 0.988 & 132.50 & 60.3 \\    
    MAXIM \cite{tu2022maxim} & 38.11 & 0.991 & 14.1 & 216 \\
    PMNet \cite{ye2022perceiving} & 38.41 & 0.990 & 18.90 & 81.83 \\
    DehazeFormer-M \cite{song2023vision} & 38.46 & 0.994 & 4.64 & 48.6 \\
    MB-TaylorFormer-B \cite{qiu2023mb} & 40.71 & 0.992 & 2.68 & 38.5 \\
    MITNet \cite{shen2023mutual} & 40.23 & 0.992 & 2.83 & 16.25 \\
    FOURMER \cite{zhou2023fourmer} & 37.32 & 0.990 & 1.29 & 20.6 \\
    DEA-Net \cite{chen2024dea} & 41.31 & 0.995 & 3.65 & 32.23 \\
    OKNet-S \cite{cui2024omni} & 37.59 & 0.994 & - & - \\    
    OKNet   \cite{cui2024omni} & 40.79 & \textbf{0.996} & 4.72 & 39.67 \\    
\midrule
\rowcolor{gray!20}\textbf{WaMaIR (ours)} & \textbf{41.55} & \textbf{0.996} & 6.17 & 46.11 \\
\bottomrule
\end{tabular}
\end{sc}
\end{scriptsize}
\end{table}

\begin{figure}[t]
    \centering
    \includegraphics[width=\linewidth]{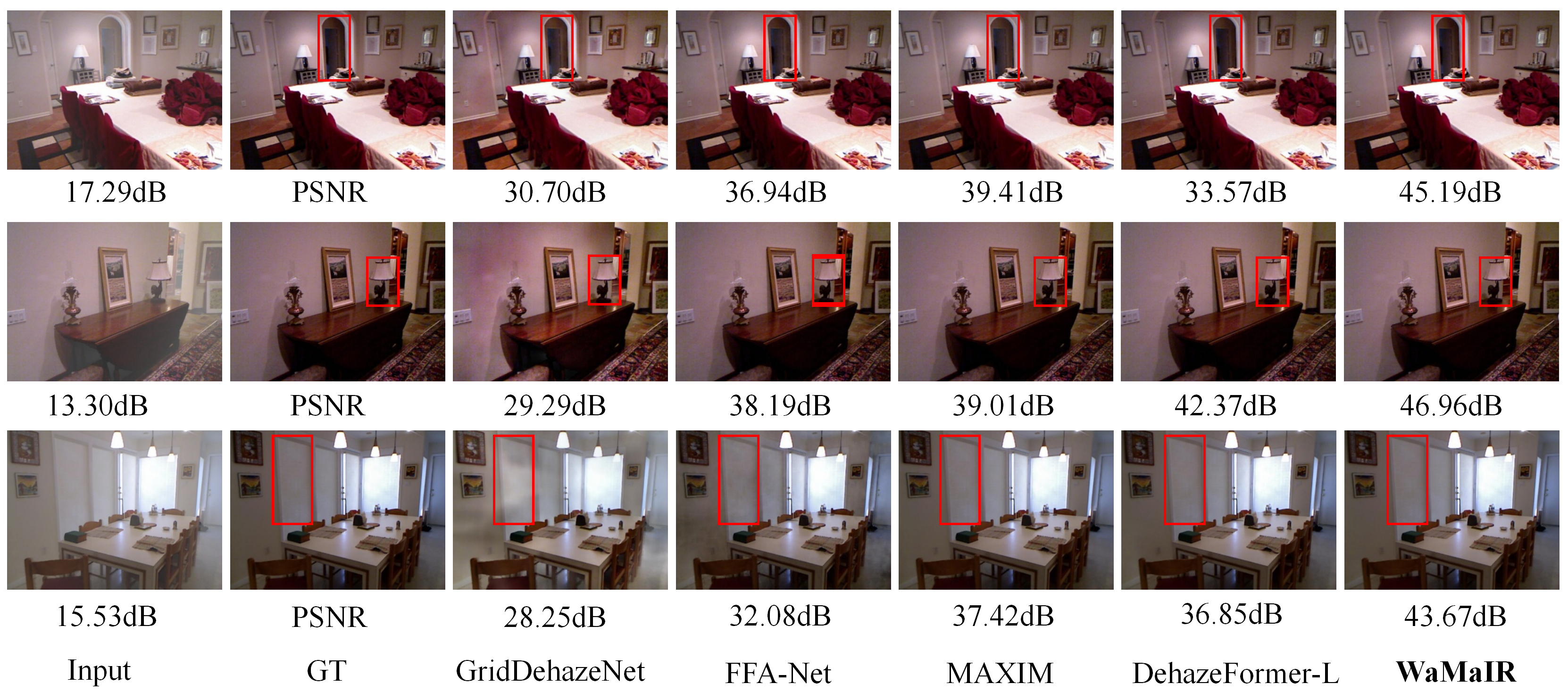}
    \caption{Image dehazing comparisions on the SOTS-indoor\cite{li2018benchmarking} dataset}
    \label{fig:dehaze}
\end{figure}
\vspace{-0.5cm}
\setlength{\textfloatsep}{10pt plus 1.5pt minus 2.5pt}

\subsection{Dehazing}

We conducted experiments on two distinct dehazing datasets: the synthetic SOTS-indoor \cite{li2018benchmarking} dataset and the nighttime NHR \cite{zhang2020nighttime} dataset. In Table~\ref{tab:reside-indoor}, our method demonstrates superior performance with a 0.76dB PSNR improvement over OKNet (a recent algorithm published within the past year) and a 0.24dB advantage over DEA-Net on the SOTS-indoor dataset. When compared to Transformer-based approaches, our method achieves a remarkable 3.09dB PSNR improvement over DehazeFormer-M while maintaining 2.46G fewer FLOPS. Additionally, it outperforms DeHamer by 4.92dB PSNR while reducing parameters to just 4.7\% of its model size. On the NHR dataset, our solution delivers a 0.47dB performance gain over the most recent state-of-the-art method in Table~\ref{tab:reside-indoor} and Table~\ref{tab:NHR}.
The results of the dehazing qualitative experiments are shown in Figure~\ref{fig:dehaze}. Our method effectively preserves texture features and achieves superior visual quality, as shown in the Figure~\ref{fig:dehaze}.The enhanced images demonstrate remarkable clarity in fine details while maintaining natural color fidelity, particularly in handling challenging haze conditions. Comparative results clearly show that our approach outperforms existing methods in both structural preservation and perceptual quality, especially in scenes with dense atmospheric interference.

\begin{minipage}{\textwidth}
\begin{minipage}{0.37\textwidth}
\makeatletter\def\@captype{table}
\caption{NightTime Image Dehazing Comparisons on the NHR \cite{zhang2020nighttime} Dataset}
\begin{scriptsize}
\begin{tabular}{l|cc}
\toprule
    \textbf{Methods} & \textbf{PSNR} & \textbf{SSIM}  \\
    \midrule
    MRP \cite{zhang2017fast}      & 19.93 & 0.777   \\
    OSFD \cite{zhang2020nighttime}           & 21.32 & 0.804  \\
    HCD \cite{wang2024restoring}     & 23.43 & 0.953   \\
    FocalNet \cite{cui2023focal}       & 25.35 & 0.969   \\
    Jin et al \cite{jin2023enhancing}         & 26.56 & 0.890  \\
    OKNet \cite{cui2024omni}   & 27.92 & 0.979   \\
    ConvIR-S \cite{cui2024revitalizing} & 28.85 & 0.981 \\
\midrule
\rowcolor{gray!20} \textbf{WaMaIR(ours)} & \textbf{28.93} & \textbf{0.982}  \\ 
\bottomrule
\end{tabular}%
\label{tab:NHR}
\end{scriptsize}
\end{minipage}
\hspace{0.02\textwidth}
\begin{minipage}{0.48\textwidth}
\makeatletter\def\@captype{table}
\begin{scriptsize}
\caption{Deraining Test100 \cite{zhang2019image} and Test2800 \cite{fu2017removing}}
\begin{tabular}{l|cc|cc}
\toprule
 & \multicolumn{2}{c|}{\textbf{Test100}} & \multicolumn{2}{c}{\textbf{Test2800}} \\
\cmidrule(lr){2-3} \cmidrule(lr){4-5}
\multirow{-2}{*}{\textbf{Methods}} & \textbf{PSNR} & \textbf{SSIM} & \textbf{PSNR} & \textbf{SSIM} \\
\midrule
    UMRL \cite{yasarla2019uncertainty}            & 24.41 & 0.829 & 29.97 & 0.905      \\
    RESCAN \cite{li2018recurrent}           & 25.00 & 0.835 & 31.29 & 0.904   \\
    PreNet \cite{ren2019progressive}           & 24.81 & 0.851 & 31.75 & 0.916  \\
    MSPFN \cite{jiang2020multi}            & 27.50 & 0.876 & 32.82 & 0.930   \\
    MPRNet \cite{zamir2021multi}           & 30.27 & 0.897 & 33.64 & 0.938   \\
    FSNet \cite{cui2023image}           & \textbf{31.05} & 0.919 & 33.64 & 0.936  \\
\midrule
\rowcolor{gray!20} \textbf{WaMaIR(ours)} & 30.00 & \textbf{0.935} & 32.35 & \textbf{0.964}  \\ 
\bottomrule
\end{tabular}
\label{tab:deraining}
\end{scriptsize}
\end{minipage}
\end{minipage}

\subsection{Deraining}

\begin{figure}[t!]
    \centering
    \includegraphics[width=\linewidth]{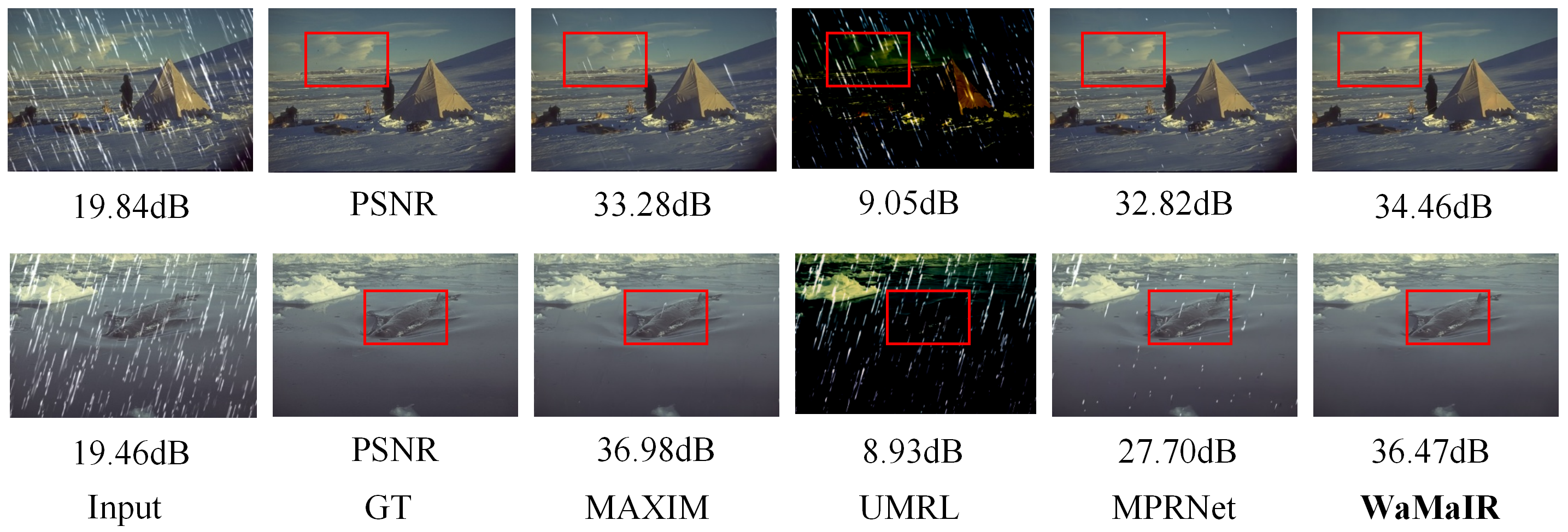}
    \caption{Image deraining comparisons on the Test100 \cite{zhang2019image} dataset.}
    \label{fig:derain}
\end{figure}

We trained our model on a composite dataset (Rain14000 \cite{fu2017removing}, Rain1800 \cite{yang2017deep}, Rain800 \cite{zhang2019image}, Rain12 \cite{li2016rain}) and evaluated it on the test sets Test100 \cite{zhang2019image} and Test2800 \cite{fu2017removing}, with the results presented in Table~\ref{tab:deraining}. In quantitative comparisons with other methods, our approach achieves excellent performance among recent techniques, obtaining the best SSIM scores on both Test100 and Test2800, while maintaining comparable PSNR results to state-of-the-art methods published within the past year.
The results of the deraining qualitative experiments are shown in Figure~\ref{fig:derain}. As can be observed, our method effectively removes rain streaks while preserving finer image details and textures. Compared to other approaches, our model demonstrates superior performance in handling heavy rain conditions and complex background scenes.

\setlength{\textfloatsep}{10pt plus 2.0pt minus 3.0pt}

\begin{figure}[h]
    \centering
    \includegraphics[width=\linewidth]{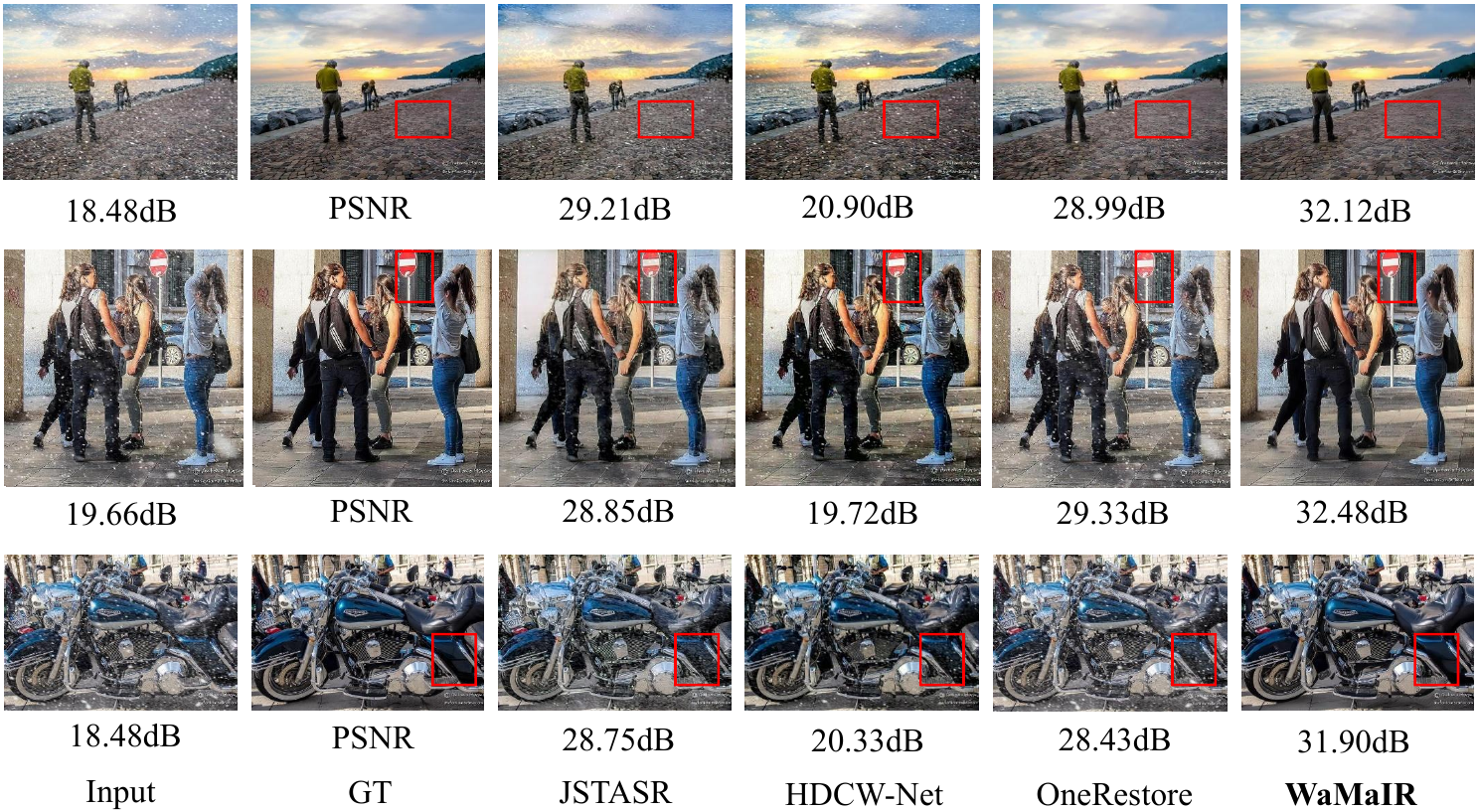}
    \caption{Image desnowing comparisons on the Snow100K \cite{liu2018desnownet} dataset.}
    \label{fig:desnow}
\end{figure}

\subsection{Desnowing}

We conducted performance comparison tests on the benchmark dataset Snow100K, as shown in Table \ref{tab:snow100k}. Our method significantly outperforms existing approaches, achieving a 0.3dB PSNR improvement over the recent Focal-Net, 0.22dB over IRNext-B, and 0.04dB over ConvIR-S. 
The results of the desnowing qualitative experiments are shown in Figure~\ref{fig:desnow}. This precise separation enables exceptionally clean restoration while preventing common artifacts like over-smoothing or residual snow patterns seen in other methods. Notably, our approach maintains consistent performance across varying snow intensities from light snowfall to blizzard conditions where most comparative methods show significant degradation.

\begin{table}[!h]
\renewcommand{\arraystretch}{0.97}
\setlength{\tabcolsep}{5pt}
\caption{Quantitative experiment on Snow100K \cite{liu2018desnownet}}
\label{tab:snow100k}
\centering
\begin{scriptsize}
\begin{sc}
\begin{tabular}{l|cc|c|c}
\toprule
    \textbf{Methods} & \textbf{PSNR} & \textbf{SSIM} & \textbf{Params/M} & \textbf{FLOPs/G} \\
    \midrule
    JSTARSR \cite{chen2020jstasr}          & 23.12 & 0.86 & 65    & \textbf{--}      \\
    HDCW-Net \cite{chen2021all}         & 31.54 & 0.95 & 6.99  & 9.78   \\
    SMGARN \cite{zhang2019image}           & 31.92 & 0.93 & 6.86  & 450.3  \\
    TransWeather \cite{valanarasu2022transweather}     & 31.82 & 0.93 & 21.9  & 5.64   \\
    FocalNet \cite{cui2023focal}         & 33.53 & 0.95 & 3.74  & 30.63  \\
    IRNext-B \cite{cui2023irnext}           & 33.61 & 0.95 & 5.46  & 42.09  \\
    ConvIR-S \cite{cui2024revitalizing}     & 33.79 & 0.95 & 5.53  & 42.1  \\
\midrule
\rowcolor{gray!20} \textbf{WaMaIR(ours)} & \textbf{33.83} & \textbf{0.95} & 6.17  & 46.11  \\ 
\bottomrule
\end{tabular}
\end{sc}
\end{scriptsize}
\end{table}

\begin{minipage}{\textwidth}
\begin{minipage}{0.4\textwidth}
\makeatletter\def\@captype{table}
\caption{Break-down ablation studies for GMWTConvs and MCAM.}
\begin{scriptsize}
\begin{tabular}{l|cccc}
\toprule
    Methods & a & b & c & d \\
    \midrule
    baseline        & \checkmark & \checkmark & \checkmark  & \checkmark   \\
    GMWTConvs       &  & \checkmark &     & \checkmark  \\
    MCAM         &  &  & \checkmark    & \checkmark      \\
\midrule
 PSNR & 28.24 & 28.29 & 28.67  & 28.93  \\ 
 Params/M & 5.53 & 5.61 & 6.1  & 6.17  \\ 
 FLOPs/G & 42.23 & 42.46 & 45.89  & 46.11  \\ 
 Times/s & 0.033 & 0.049 & 0.044  & 0.058  \\ 
\bottomrule
\end{tabular}%
\label{tab:ablation1}
\end{scriptsize}
\end{minipage}
\hspace{0.1\textwidth}
\begin{minipage}{0.32\textwidth}
\makeatletter\def\@captype{table}
\begin{scriptsize}
\caption{Break-down ablation studies for loss function.}
\begin{tabular}{l|ccc}
\toprule
    Methods & a & b & c  \\
    \midrule
    spartial loss        & \checkmark & \checkmark & \checkmark     \\
    frequecy loss       &  & \checkmark &  \checkmark  \\
    wavelet loss          &  &      & \checkmark      \\
\midrule
 PSNR & 28.48 & 28.63 & 28.93    \\ 
 SSIM & 0.980 & 0.981 & 0.982    \\ 
\bottomrule
\end{tabular}%
\label{tab:ablation2}
\end{scriptsize}
\end{minipage}
\end{minipage}

\subsection{Ablation Study}

We conducted ablation studies on the NHR \cite{zhang2020nighttime} dataset with hyperparameter n=3 and 300 training epochs. The baseline was obtained by removing both GMWTConvs and MCAM.

\textbf{Component-wise ablation.} The progressive module integration tests revealed in Table~\ref{tab:ablation1} that the baseline initially achieved 28.24dB PSNR, with subsequent additions of GMWTConvs and MCAM yielding improvements of 0.05dB and 0.43dB respectively, while our complete model demonstrated 0.69dB enhancement over the baseline with modest computational overhead - merely 0.64M additional parameters, 3.88G extra FLOPS, and only 0.025s increased processing time per image. 

\textbf{Loss function comparison.} In parallel loss function experiments documented in Table~\ref{tab:ablation2}, the spatial-loss-only baseline scored 28.48dB PSNR, with the incorporation of frequency spectrum loss ($\lambda$=0.1) and wavelet domain loss ($\lambda$=0.05) contributing 0.15dB improvement and ultimately delivering our final model's 0.45dB superiority over the baseline configuration.

\section{Conclusion}

In this paper, we introduced WaMaIR, a novel image restoration framework designed specifically to enhance texture reconstruction by effectively expanding receptive fields and capturing detailed feature dependencies. We proposed the Global Multiscale Wavelet Transform Convolutions (GMWTConvs), significantly enlarging receptive fields and enriching multiscale texture features. To better capture channel-wise dependencies and enhance sensitivity to color, edges, and textures, we introduced the Mamba-Based Channel-Aware Module (MCAM). Additionally, we developed a Multiscale Texture Enhancement Loss (MTELoss) to explicitly guide the model towards accurately preserving intricate texture details. Comprehensive experiments demonstrated that WaMaIR outperforms existing state-of-the-art methods, achieving superior image restoration performance while maintaining computational efficiency.

\bibliographystyle{splncs04}
\bibliography{main}

\end{document}